\documentclass{article}
\usepackage{amsmath,epsfig}
\usepackage{amssymb,multirow}
\usepackage[ruled]{algorithm2e}
\usepackage[T1]{fontenc}

\usepackage[preprint]{spconfa4}
\let\OLDthebibliography\thebibliography
\renewcommand\thebibliography[1]{
	\OLDthebibliography{#1}
	\setlength{\parskip}{0pt}
	\setlength{\itemsep}{0pt plus 0.3ex}
}

\usepackage{fancyhdr}

\begin{document}\sloppy
	
	% Example definitions.
	% --------------------
	\def\x{{\mathbf x}}
	\def\L{{\cal L}}

	% Title.
	% ------
	\title{Spatial-Temporal Human-Object Interaction Detection}
	%
	% Single address.
	% ---------------
	% \name{Anonymous ICME submission}
	%Address and e-mail should NOT be added in the submission paper. They should be present only in the camera ready paper. 
	\name{
		Xu Sun$^{1,2}$, Yunqing He$^{1}$, Tongwei Ren$^{1,2,*}$, Gangshan Wu$^{1}$
	}
	% \address{%ss
	% 	$^{1}$State Key Laboratory for Novel Software Technology, Nanjing University, Nanjing, China
	% }
	% \address{%ss{%
	% 	$^{2}$Shenzhen Research Institute of Nanjing University, Nanjing University, Shenzhen, China
	% }
	% \address{{sunx, yqhe}@smail.nju.edu.cn, {rentw, gswu}@nju.edu.cn}
	\address{
		$^{1}$State Key Laboratory for Novel Software Technology, Nanjing University, Nanjing, China \\
		$^{2}$Shenzhen Research Institute of Nanjing University, Nanjing University, Shenzhen, China \\
		\{sunx, yqhe\}@smail.nju.edu.cn, \{rentw, gswu\}@nju.edu.cn
	}
	
	\maketitle
	
	\thispagestyle{fancy}
	\fancyhead{}
	\lhead{}
	% \lfoot{978-1-6654-3864-3/21/\$31.00~\copyright~2021 IEEE}
	\cfoot{}
	\rfoot{}

	\begin{abstract}
		In this paper, we propose a new instance-level human-object interaction detection task on videos called ST-HOID, which aims to distinguish fine-grained human-object interactions (HOIs) and the trajectories of subjects and objects.
		It is motivated by the fact that HOI is crucial for human-centric video content understanding.
		To solve ST-HOID, we propose a novel method consisting of an object trajectory detection module and an interaction reasoning module.
		Furthermore, we construct the first dataset named VidOR-HOID for ST-HOID evaluation, which contains 10,831 spatial-temporal HOI instances.
		We conduct extensive experiments to evaluate the effectiveness of our method.
		The experimental results demonstrate that our method outperforms the baselines generated by the state-of-the-art methods of image human-object interaction detection, video visual relation detection and video human-object interaction recognition.
	\end{abstract}
	\begin{keywords}
		Human-object interaction detection, spatial-temporal human-object interaction, object trajectory generation, dynamic interaction recognition
	\end{keywords}

	\section{Introduction}
	\label{sec:intro}
	
	Human-object interaction (HOI) describes fine-grained interactions between humans and objects~\cite{chao2015hico}.
	It is crucial for human-centric visual content understanding, because accurate HOI detection (HOID) results play a fundamental role in numerous multimedia applications, such as captioning~\cite{dong2016early}, multi-modal dialog~\cite{liao2018knowledge} and visual question answering~\cite{antol2015vqa}.
	
	Most existing HOID methods are proposed for still images (ImgHOID)~\cite{icme20imghoid1, icme20imghoid2}.
	These methods perform poor when directly applying them for HOID on videos (VidHOID), because they cannot effectively explore temporal cues from videos.
	For one thing, HOIs in videos are usually changeable over time,~\emph{e.g.}, they may be not detected because humans/objects are occluded.
	For another, VidHOID requires to localize humans and objects with continuous object trajectories across video frames instead of independent bounding boxes on different frames.
	Meanwhile, VidHOID is more natural to distinguish fine-grained HOIs by exploring dynamic content,~\emph{e.g.}, ``wave'' and ``hold'' have similar still appearances and they cannot be effectively distinguished on images.
	
	Though there have been some works for HOI analysis on videos, they omit some important characteristics of VidHOID, such as humans/objects localization~\cite{baradel2018object}.
	There are also some video analysis tasks similar to VidHOID.
	For example, spatial-temporal action detection (STAD)~\cite{ST_AR} extracts the trajectories of individuals and recognize the human actions, but it ignores the objects being interacted with and it assumes that there is only one object is being interacting with simultaneously; video visual relation detection (VidVRD)~\cite{cvpr20vidvrd} aims to generate holistic description for video content with spatial-temporal relation instances, including visual relations in the format of $\langle$subject, predicate, object$\rangle$ and the trajectories of subjects and objects, but the detected visual relation instances are not human-centric.
	
	\begin{figure}[!t]
		\begin{center}
			\includegraphics[width=\linewidth]{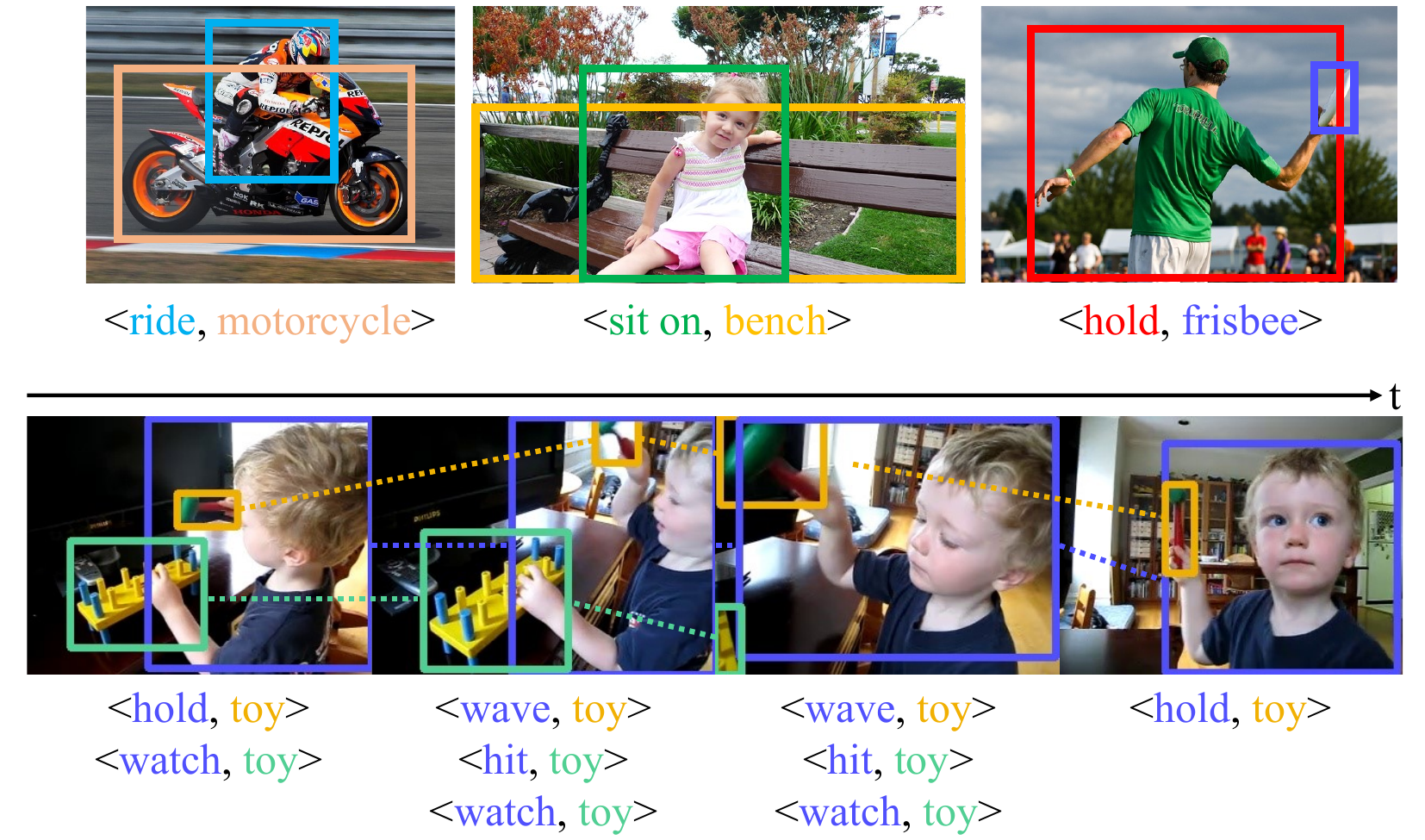}
		\end{center}
		\vspace{-0.7cm}
		\caption{
			Comparison of ImgHOID and ST-HOID.
			The top row shows three ImgHOID examples represented with HOI labels and localized human-object pairs.
			The bottom row shows a ST-HOID example using trajectories to indicate the locations of entities and temporally-consistent label sequences to indicate the dynamic HOIs.
			The colors show the consistence between the labels and bounding boxes.
		}
		\label{fig:fig_intro}
	\vspace{-0.5cm}
	\end{figure}
	
	To this end, we propose a new HOID task on videos, which provides instance-level VidHOID by taking untrimmed videos as input and generating a set of spatial-temporal HOIs consisting of a predicate and object pair $\langle$predicate, object$\rangle$ and the corresponding trajectories of the subject and object.
	To discriminate the new task with the existing works, we name it as Spatial-Temporal Human-Object Interaction Detection (ST-HOID).
	Compared with unhuman-centric tasks, i.e., VidVRD, ST-HOID focus on more meaningful information for deep scene understanding other than common visual facts like $\langle$vase, on, desk$\rangle$.
	Figure~\ref{fig:fig_intro} shows the differences between ImgHOID and ST-HOID, which demonstrates advantages of dynamic HOID to static HOID.
	
	To the best of our knowledge, there is no existing dataset suitable for ST-HOID evaluation.
	Thus, we construct the first ST-HOID dataset VidOR-HOID based on the large-scale  dataset VidOR~\cite{shang2019annotating}, which contains 1,134 videos with densely annotated human and object trajectories and instance-level HOI labels.
	Compared with previous datasets for VRD task, we exclude the predicates such as ``on'' and ``in'', because they are usually too general to represent the fine-grained interactions between humans and objects.
	
	The main contributions of this paper are threefold:	
	(1) A new VidHOID task named ST-HOID, which aims to generate instance-level HOIs for untrimmed videos.
	(2) A novel ST-HOID method consisting of an object trajectory detection module based on object detection and visual tracking, and an interaction reasoning module based on multi-modal feature fusion.
	(3) The first ST-HOID dataset VidOR-HOID that contains 1,134 videos from VidOR dataset.
	
	\section{Related Work}
	
	ImgHOID has drawn much attention from multimedia and computer vision communities~\cite{chao2018learning}.
	Shen~\emph{et al.} explore zero-shot learning and effectively scale ImgHOID~\cite{shen2018scaling}.
	Xu~\emph{et al.} investigate intrinsic semantic regularities from visual and linguistic information with multi-modal feature embedding and satisfying performance is achieved ~\cite{xu2019learning}.
	An effective model, iCAN, is proposed by Gao~\emph{et al.} to enhance ImgHOID with instance-centric attention mechanism~\cite{gao2018ican}.
	However, existing ImgHOID methods can only utilize the static features, which are insufficient to recognize the dynamic HOIs in videos.
	
	VidHOIR aims to recognize the human-object interactions on video frames or segments disregard of the corresponding human and object instances~\cite{qi2018learning}.
	Zhou~\emph{et al.} design a general structure to explore the common visual cues at multiple time scales, and prove that the models equipped with the temporal relational network can effectively recognize HOIs in videos~\cite{zhou2018temporal}.
	However, VidHOIR only focuses on HOI tagging while ST-HOID requires not only HOI recognition but also spatial-temporal localization, which provides understanding in depth to complex scenes with multiple interactive entities.

	VidVRD aims to capture spatial-temporal relation instances in videos, which are represented with relation triplets and object trajectory-pairs~\cite{shang2017video}.
	Recently, another large-scale VidVRD dataset, VidOR, is constructed, which serves as a new benchmark towards daily video understanding~\cite{shang2019annotating}.
	Sun~\emph{et al}. explicitly combine the relative motion feature and the language context feature to perform visual relation detection and achieve state-of-the-art performance~\cite{sun2019video}.
	Compared with ST-HOID, VidVRD focuses on salient objective relations in holistic video, but latent interactions between humans could be omitted, which is essential for some downstream tasks like social relation analysis.
	
	STAD focuses on the dynamic human behavior in videos, which aims to localize the individuals and recognize the actions simultaneously.
	AVA~\cite{gu2018ava} is one of the most frequently-used datasets for spatial-temporal action detection evaluation.
	Singh~\emph{et al.} present a real-time action detection model by extending an online object detection framework~\cite{singh2017online}.
	% Later related works also consider classless objects detection, but only as supplementary information.
	Sun~\emph{et al.} emphasize the importance of spatial-temporal context information for action detection and propose a weakly supervised actor-centered relation network to investigate the influence of context~\cite{sun2018actor}.
	Spatial-temporal action detection only concentrates on the human actions without explicitly considering the context, especially when a certain person is interacting with multiple objects simultaneously, which conveys less semantic information than ST-HOID.

	\section{Method}
	
	\begin{figure*}[!th]
		\begin{center}
			\includegraphics[width=0.85\linewidth]{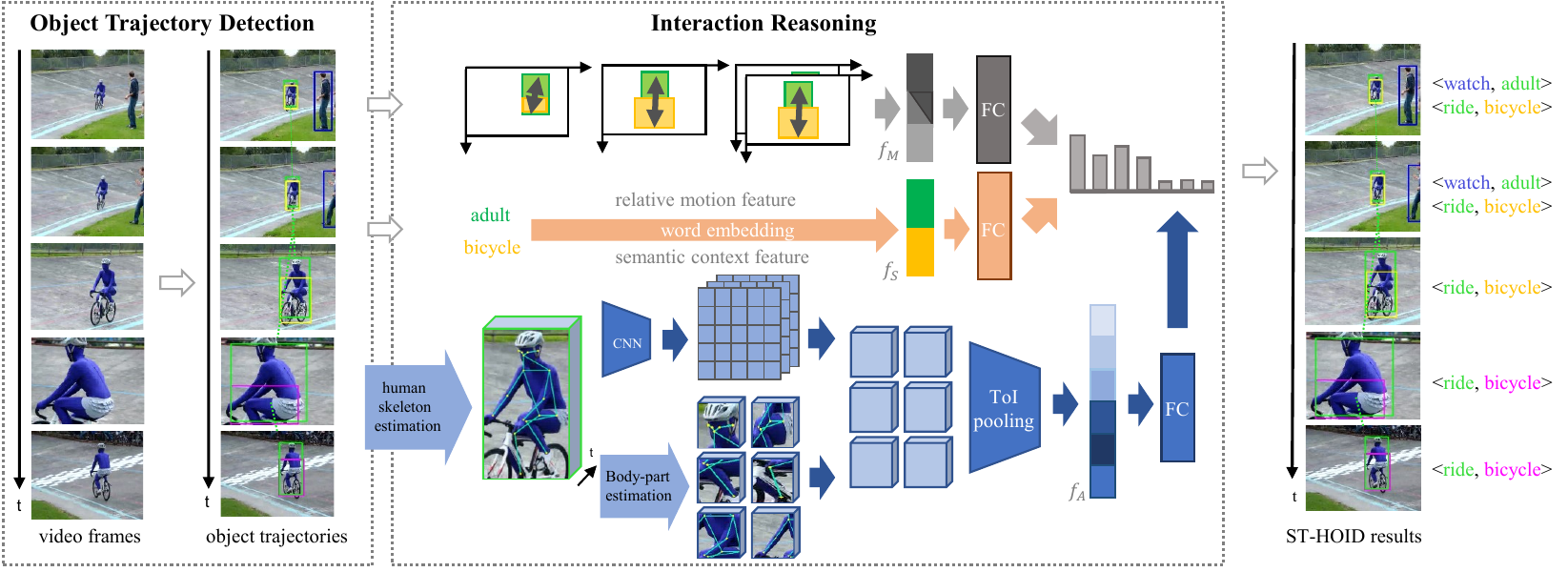}
		\end{center}
		\vspace{-0.7cm}
		\caption{An overview of the proposed ST-HOID method.}
		\label{fig:fig-method}
		\vspace{-0.5cm}
	\end{figure*}
	
	\label{sec:method}
	
	\subsection{Object trajectory detection}
	We propose an effective object trajectory detection module to localize objects with temporally-consistent bounding box sequences.
	Inspired by existing VidVRD methods~\cite{shang2017video, sun2019video}, we split a complete video into a series of temporally-overlapping segments with fixed duration and greedily merge the obtained short-term ones extracted from video segments.
	
	Short-term trajectories on each video segment are generated by using visual tracker initialized with frame-level object detection results.
	Considering that existing visual tracking methods requires great computation resource, and often effected by severe occlusion and motion blur, we propose a novel short-term trajectory generation algorithm.
	For each video segment, we collect and sort the detected objects on all frames according to their prediction confidences, which are referred to as $\mathcal{O}{=}\{(b_{f,m},c_{f,m},s_{f,m})|f{=}1,...,L\}$.
	Here, $b$, $c$ and $s$ denote the bounding box, category and confidence of a detected object on a frame, respectively; $f$ and $m$ denote the frame index in segment and the detection index on frame.
	The frame-level detection with the highest confidence is the first to be tracked and extended to a trajectory.
	Then we calculate IoUs (intersection over union) between the untracked objects $O_u$ and the bounding boxes of the newly-generated trajectory, $\hat{\mathcal{T}}{=}\{b_f|f{=}1,...,L\}$.
	The greatly-overlapping frame-level detection are removed from $O_u$ and merged into $\hat{\mathcal{T}}$.
	This procedure is repeated until $O_u$ is empty.
	Assume
	\begin{small}
		$\hat{\mathcal{T}}_x{=}\{b_f|f{=}m,...,n\}$
	\end{small}
	and
	\begin{small}
		$\hat{\mathcal{T}}_y{=}\{b_f|f{=}p,...,q\}$
	\end{small}
	are two short-term trajectories with the same object category extracted from two temporally-overlapping video segments. 
	The trajectory overlap ratio $\Theta$ between $\hat{\mathcal{T}_x}$ and $\hat{\mathcal{T}_y}$ is calculated as Eq.~(\ref{equ:OR}), where $\vartheta(\cdot)$ is a function, which returns 1 if the condition inside is true, otherwise returns 0; $\beta$ equals 0.5 in our experiments.
	\begin{equation}
	\label{equ:OR}
	\Theta(\hat{\mathcal{T}}_x, \hat{\mathcal{T}}_y)\!=\!
	\begin{cases}
	\frac{\sum\limits_{i=\max(m, p)}^{\min(n, q)}\vartheta(\text{IoU}(b_{i,x}, b_{i,y}) > \beta)}{\min(n, q) - \max(m, p) + 1}, &n \!\ge\! p \ \text{and} \ m \!\le\! q \\
	0, &\text{otherwise}
	\end{cases}.
	\setlength\abovedisplayskip{3pt}
	\setlength\belowdisplayskip{3pt}
	\end{equation}
	
	\subsection{Interaction reasoning}
	We combinatorially pair the human trajectories and object trajectories and extract their co-occurrent parts as HOI candidates $\mathcal{P}{=}(\mathcal{T}_h, \mathcal{T}_o)$.
	$\mathcal{P}$ is split into consecutive short-term segments $\mathcal{S}{=}\{(\hat{\mathcal{T}}_h^i, \hat{\mathcal{T}}_o^i)\}$, whose durations are fixed (\emph{e.g.}, $|\hat{\mathcal{T}}^i_h|{=}|\hat{\mathcal{T}}^i_o|{=}10$), which serve as the atomic elements for interaction recognition.
	The consecutive candidate segments recognized as the same interaction category are greedily associated to complete spatial-temporal HOI instances.
	To understand the fine-grained interactions between a human and an object, comprehensive features are necessary to provide the information of both human behavior and object context.
	In this paper, we propose an interaction recognition module using multi-modal feature fusion.
	      
	\textbf{Human behavior descriptor}. To capture the fine-grained human behavior, we focus on the motion of human body parts.
	We apply an off-the-shelf multi-person pose estimation method, RMPE~\cite{fang2017rmpe}, which is pretrained on MSCOCO-keypoint dataset, to obtain the body structure information.
	It represents a human instance on a video frame with a skeleton $\mathcal{H}{=}\{p_1,...,p_{N_k}\}$, where $p_i$ denotes the coordinate of the $i$-th body joint and $N_k{=}17$.
	We assign the obtained frame-level skeletons to the corresponding human trajectories according to the IoUs between the bounding boxes of the skeletons and trajectories on each frame.
	In this way, the skeletons obtained from individual frames are associated to temporally-consistent skeleton trajectories $\mathcal{T}'_h=\{\mathcal{H}_f\}$.

	Considering the fact that an interaction is performed by different body parts jointly, we extract and combine the dynamic visual features of different body parts for fine-grained HOI reasoning.
	Specifically, for an HOI candidate segment $(\hat{\mathcal{T}}_h, \hat{\mathcal{T}}_o)$, $\hat{\mathcal{T}}_h$ can be separated into $N_k$ body part trajectory segments $\{\hat{\mathcal{T}}^1_p, ..., \hat{\mathcal{T}}^{N_k}_p\}$ as mentioned above.
	For each $\hat{\mathcal{T}}_p$, we apply ToI-pooling (trajectory of interest)~\cite{hou2017tube} to encode its dynamic appearance information.
	ToI-pooling takes $\hat{\mathcal{T}}$ and CNN feature maps extracted from the video segment containing $\hat{\mathcal{T}}$ as inputs.
	According to each bounding box $b_i$ of $\hat{\mathcal{T}}$, frame-level feature $f_i$ is obtained by applying RoI-align on the $i$-th CNN feature map.
	We fuse all the frame-level features $\{f_1, f_2, ..., f_{|\hat{\mathcal{T}}|}\}$ extracted from $\hat{\mathcal{T}}$ to generate a dynamic visual feature $\tilde{f}$ for a body part trajectory segment as following:
	\begin{equation}
	\label{equ:fuse}
	\tilde{f} = \mathcal{M}(\{f_1, f_2, ..., f_{|\hat{\mathcal{T}}|}\}),
	\setlength\abovedisplayskip{3pt}
	\setlength\belowdisplayskip{3pt}
	\end{equation}
	where $\mathcal{M}(\cdot)$ denotes element-wise maximum.
	Then we obtain a feature vector $v$, whose length is $c$, by applying global-average-pooling on $\tilde{f}$.
	For one thing, ToI-pooling reserves the spatial information of 2D feature maps.
	For another, ToI-pooling can tolerate trajectories with different sizes.
	We combine the feature vectors extracted from $N_p$ body part trajectory segments as the human behavior descriptor:
	\begin{equation}
	\label{equ:combine}
	f_{A} = \biguplus\limits_{k=1,...,N_p} v_k,
	\setlength\abovedisplayskip{3pt}
	\setlength\belowdisplayskip{3pt}
	\end{equation}
	where $v_k$ corresponds to the $k$-th body part and $\biguplus$ denotes accumulated feature concatenation.

	\textbf{Relative motion feature and semantic context feature}.
	Besides the dynamic visual feature of human body parts, we further include a relative motion feature $f_{M}$ and semantic context feature $f_{C}$ to enhance the capability of the interaction recognition module.
	For an HOI candidate segment $(\hat{\mathcal{T}}_h, \hat{\mathcal{T}}_o)$, $f_{M}$ encodes the relative location and motion between the human and object over time, the effectiveness of which has been validated by video visual relation detection method VidVRD-MMF~\cite{sun2019video}.
	Specifically, the relative location is encoded as $f^t_{LOC}{=}(s_x, s_y, s_w, s_h, s_a)$.
	Here, $s_x{=}\frac{x - x'}{w}$, $s_y{=}\frac{y - y'}{h}$, $s_w{=}\log\frac{w}{w'}$, $s_h{=}\log\frac{h}{h'}$, $s_a{=}\log\frac{w \cdot h}{w' \cdot h'}$, $(x,y,w,h)$ and $(x',y',w',h')$ are the $t$-th bounding boxes of $\hat{\mathcal{T}}_h$ and $\hat{\mathcal{T}}_o$, respectively.
	The relative motion feature $f_{M}$ is calculated as following:
	\begin{equation}
	\label{equ:motion}
	f_{M} = f^1_{LOC} \oplus f^L_{LOC} \oplus (f^{L}_{LOC} - f^1_{LOC}),
	\setlength\abovedisplayskip{3pt}
	\setlength\belowdisplayskip{3pt}
	\end{equation}
	where $\oplus$ denotes feature concatenation; $L{=}|\hat{\mathcal{T}}_h|{=}|\hat{\mathcal{T}}_o|$,~\emph{i.e.}, the temporal length of HOI candidate segment.
	Besides, considering that certain interactions are closely related to certain human and object categories, we utilize a set of word vectors pretrained on large-scale linguistic dataset to represent different human and object categories following VidVRD-MMF.
	We concatenate the two word vectors $v_h$ and $v_o$ corresponding to the categories of $\hat{\mathcal{T}}_h$ and $\hat{\mathcal{T}}_o$ as semantic context feature $f_{S}{=}v_h \oplus v_o$.
	
	\textbf{Factorized interaction recognition}. 
	Considering the great number of interaction category combinations, we decompose interaction recognition into object classification within an object category set $\Omega$ and predicate classification within a predicate category set $\Phi$.
	The object category $\omega\in\Omega$ is recognized in object trajectory detection and the predicate classification relies on the object category $\omega$ and obtained multi-modal features $f_{A}$, $f_{M}$ and $f_{S}$, which are fed into three independent classifiers $\mathcal{C}$ to estimate the probabilities of all predicate categories,
	where the subscript $t{\in}\{A,M,S\}$ denotes the type of feature and $\lambda_\omega$ is a hard attention mask.
	Specifically, $\lambda_\omega$ is a binary vector containing $|\Phi|$ elements, each of which corresponds to a certain predicate of $\Phi$.
	If a predicate is never related to an object categorized as $\omega$ in the training data, the corresponding element is set to zero and otherwise one.
	In this way, the easy-negative predicate categories are filtered out by the interacted object context, which also makes training focuses on the related predicates.
	
	\textbf{Training loss}. 
	Since different predicates are not mutually exclusive, ~\emph{e.g.}, ``watch'' and ``cut'' a ``cake'', we assume that different interactions are performed independently and utilize binary cross entropy loss for training.
	During training, a data sample is a manually annotated HOI segment ($\hat{\mathcal{T}}^g_h$, $\hat{\mathcal{T}}^g_o$, $\langle \varphi_g, \omega_g \rangle$).
	The predicate category $\omega_g$ is represented with a binary vector $\gamma_g$, whose length is equal to $|\Phi|$, and the prediction of $\gamma_g$ is represented as a vector $\hat{\rho}$. 
	The element of $\gamma_g$ corresponding to $\varphi_g$ is set to one and others are set to zero.
	The training loss is calculated as following:
	\begin{equation}
	\vspace{-0.5cm}
	\begin{aligned}
	\label{equ:loss}
	\setlength\abovedisplayskip{6pt}
	\mathcal{L} &= -\sum_{t\in[A, M, S]}{\sum_{i=1}^{|\Phi|}{\gamma^i_g\log(\hat\rho^i_t)+(1-\gamma^i_g)\log(1-\hat\rho^i_t)}}.
	\setlength\belowdisplayskip{3pt}
	\end{aligned}
	\end{equation}
	
	\section{Experiments}
	
	\subsection{Metrics and dataset}
	We inherit class mAP from ImgHOID as a primary metric,~\emph{i.e.}, mean average precision over classes, which is widely used in detection tasks.
	It can effectively validate the generalizing-capability of methods and is sensitive to the models that over fit the frequency of data.
	Following VidVRD, we adopt video mAP as well,~\emph{i.e.}, mean average precision over videos.
	Considering the fact that incomplete annotation is an inevitable problem in constructing datasets for complex semantic analysis tasks such as ST-HOID and VidVRD, we also adopt recall@K following VidVRD.
	
	Owing to the fact that object trajectory detection is still an open problem which is hardly explored, we evaluate the proposed method on a task named HOI tagging.
	It takes videos as input and predicts a set of HOI labels as output, eliminating the influence of object trajectory detection and is compatible to existing VidHOIR methods.
	We utilize precision@N to evaluate the accuracy of predicted HOI labels following relation tagging in VidVRD, here $N$ is 1, 5 and 10.

	For ST-HOID evaluation, we randomly select 1,134 videos from VidOR, in which the spatial-temporal HOI instances are reserved while others are eliminated.
	Then the videos are split into two parts, 867 for training and 267 for testing.
	The HOI categories with less than 10 instances are removed so that the models can extract knowledge from all categories while the natural long-tailed distribution of data is preserved.
	At last, we obtain 10,831 spatial-temporal HOI instances in VidOR-HOID dataset (8,501 for training and 2,330 for testing).
	The dataset consists of 295 HOI categories with 29 predicates and 43 object categories.
	
	\subsection{Component Analysis}
	\begin{table}
		\begin{center}
			\caption{Comparison of different object trajectory detection methods on VidOR-HOID dataset.}
			\label{tab:component-otd}
			\begin{tabular}{cccccccc}
				\hline
				& mAP & FPS\\
				\hline
				IoU-match & 8.03 & 268.64 \\
				Track+tNMS & 12.23 & 2.21 \\
				Ours & 12.08 & 13.09 \\
				\hline
			\end{tabular}
		\end{center}
		\vspace{-0.7cm}
	\end{table}
	
	\begin{table*}
		\begin{center}
			\caption{Evaluation of our method with groundtruth object trajectories and different interaction recognition modules on VidOR-HOID dataset.}
			\label{tab:component}
			\begin{tabular}{cccccccccc}
				\hline
				\multirow{2}{*}{Method} & \multicolumn{4}{c}{ST-HOID} & & & \multicolumn{3}{c}{HOI tagging}\\
				\cline{2-10}
				& class mAP & video mAP & R@50 & R@100 & & & P@1 & P@5 & P@10\\
				\hline
				Ours-GT w/o BP & 15.98 & 20.89 & 23.50 & 27.62 & & & 62.54 & 33.93 & 22.28 \\
				Ours-GT w/o OC & 10.04 & 11.93 & 20.63 & 25.20 & & & 57.30 & 31.98 & 20.49 \\
				Ours-GT w/o LF & 16.32 & 21.88 & \textbf{24.44} & \textbf{28.25} & & & 62.92 & 33.78 & 22.06 \\
				Ours-GT w/o CF & 14.78 & 18.14 & 21.52 & 24.80 & & & 59.18 & 31.76 & 22.24 \\
				Ours-GT & \textbf{16.63} & \textbf{21.95} & 23.99 & 28.07 & & & \textbf{68.53} & \textbf{34.38} & \textbf{22.40} \\
				\hline
			\end{tabular}
		\end{center}
		\vspace{-0.3cm}
	\end{table*}
	
	\begin{table*}
		\begin{center}
			\caption{Comparison of our method and the state-of-the-art baselines on VidOR-HOID dataset.}
			\label{tab:comparison}
			\begin{tabular}{cccccccccc}
				\hline
				\multirow{2}{*}{Method} & \multicolumn{4}{c}{ST-HOID} & & & \multicolumn{3}{c}{HOI tagging}\\
				\cline{2-10}
				& class mAP & video mAP & R@50 & R@100 & & & P@1 & P@5 & P@10\\
				\hline
				\multicolumn{1}{l}{STGCN~\cite{qian2019video}} & 0.29 & 0.27 & 0.66 & 0.91 & & & 26.14 & 14.13 & 10.28\\
				\multicolumn{1}{l}{VidVRD~\cite{shang2017video}} & 1.46 & 0.91 & 1.35 & 1.79 & & & 13.86 & 9.44 & 7.04 \\
				\multicolumn{1}{l}{VidVRD-MMF~\cite{sun2019video}} & 1.72 & 2.79 & 3.27 & 3.86 & & & 46.82 & 18.65 & 12.25 \\
				\multicolumn{1}{l}{TRN~\cite{zhou2018temporal}} & 0.21 & 0.01 & 0.04 & 0.09 & & & 41.95 & 19.44 & \textbf{16.38} \\
				\multicolumn{1}{l}{HO-RCNN~\cite{chao2018learning}} & 0.78 & 1.55 & 2.73 & 3.41 & & & 34.77 & 14.80 & 10.19 \\
				\multicolumn{1}{l}{iCAN~\cite{gao2018ican}} & 1.17 & 2.03 & 3.14 & 3.90 & & & 41.95 & 19.55 & 12.55 \\
				\multicolumn{1}{l}{Ours} & \textbf{2.57} & \textbf{3.22} & \textbf{3.63} & \textbf{4.13} & & & \textbf{50.94} & \textbf{20.30} & 13.07  \\
				\hline
			\end{tabular}
		\end{center}
		\vspace{-0.5cm}
	\end{table*}
	
	% \textbf{Object trajectory detection} determines the upper bound of the performance of ST-HOID methods.
	\textbf{Object trajectory detection}.
	The proposed object trajectory detection module are evaluated by comparing with two baseline methods, IoU-match and Track+TNMS.
	Specifically, IoU-match is implemented by greedily associating the frame-level detected boxes on consecutive frames.
	Track+tNMS is the object trajectory detection module adopted by VidVRD and ST-GCN, which generates short-term trajectories by tracking all frame-level detected boxes in each video segment and removing the redundant ones with tublet non-maximum-suppression (tNMS).
	The short-term trajectories extracted from consecutive video segments are greedily associated according to their overlaps, similar to our proposed method.
	To compare the effectiveness and efficiency of different methods, we adopt mAP and FPS as evaluation metrics following object detection, as shown in Table~\ref{tab:component-otd}.
	The experiments are applied on the same device with one NVIDIA GTX-1080ti GPU.
	
	\textbf{Interaction recognition}.
	Whilst object trajectory generation is still an open problem, we construct a baseline referred to as Ours-GT by replacing the detected trajectories with manual annotated ones in the proposed method to eliminate the influence of object trajectory detection.
	As shown in Table~\ref{tab:component}, we firstly construct three baselines Ours-GT w/o OC, Ours-GT w/o LF, Ours-GT w/o CF by removing the adopted three training techniques including the hard object context (OC), late feature fusion (LF) and HOI category factorization (CF), respectively.
	As shown in the last four rows in Table~\ref{tab:component}, the complete version of the proposed method achieves the best performance, demonstrating that all three training techniques adopted can improve the capability of recognition.
	Besides, we utilize a human behavior descriptor to capture the subtle motions of the individuals in video.
	To evaluate the effectiveness of it, we remove it from our method and construct a baseline Ours-GT w/o BP.
	All evaluation results of Ours-GT w/o BP are significantly lower than our method, demonstrating that adopted dynamic visual feature of human body parts can improve the capability of recognition.

	\subsection{Comparison with State-of-the-arts}
	
	\textbf{Baselines}.
	We construct two baselines by extending two ImgHOID methods, HO-RCNN~\cite{chao2018learning} and iCAN~\cite{gao2018ican}.
	Specifically, HO-RCNN and iCAN are trained using the key frames of videos.
	During testing, the obtained models are utilized to densely detect HOI instances on all video frames, which are greedily associated as spatial-temporal HOI instances.
	
	We construct three baselines based on three VidVRD methods, VidVRD~\cite{shang2017video}, STGCN~\cite{qian2019video}, and VidVRD-MMF~\cite{sun2019video}.
	We make these VidVRD methods fit ST-HOID task by filtering out the detected relation instances whose subjects are not human and eliminating the subjects from the recognized relation triplets.
	Besides, we extend a VidHOIR method TRN~\cite{zhou2018temporal}.
	Since TRN only predicts HOI labels for trimmed videos and ignores the locations of entities, we provide our detected object trajectories and take TRN as interaction reasoning head.
	We re-train these baseline methods with their provided best training strategies and pre-trained models.
	If their own strategies or model weights are not available, we adopt the same strategy as ours and standard pre-trained backbones for fair comparison.
	
	\begin{figure*}[!th]
		\begin{center}
			\includegraphics[width=0.8\linewidth]{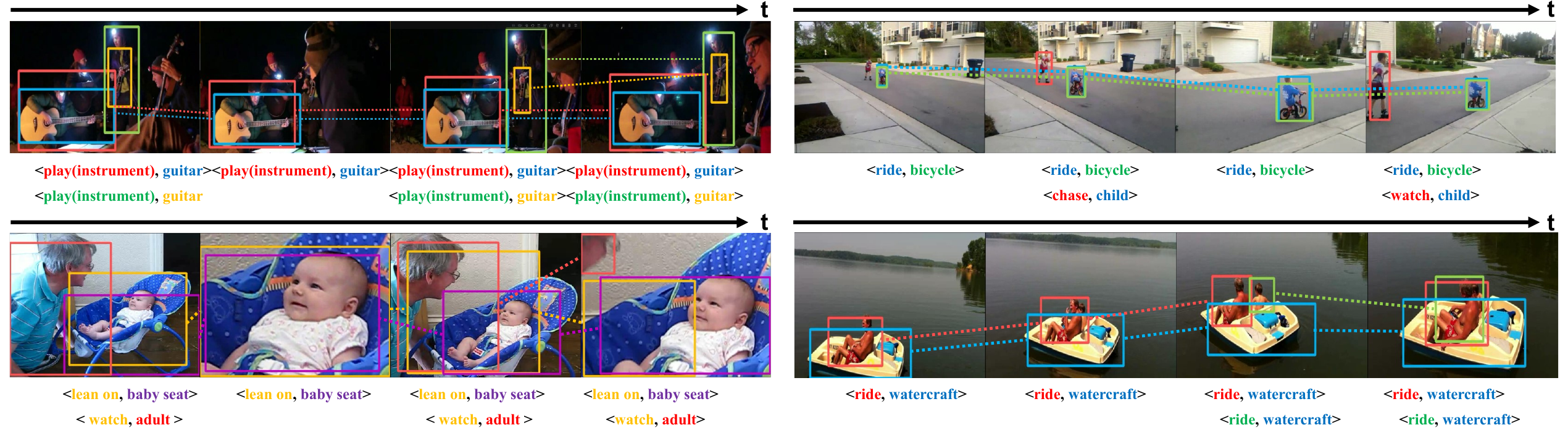}
		\end{center}
		\vspace{-0.7cm}
		\caption{Qualitative results of our method on VidOR-HOID dataset.}
		\label{fig:fig-sample}
		\vspace{-0.3cm}
	\end{figure*}
	
	\textbf{Results}.
	Table~\ref{tab:comparison} indicates the evaluation results of all comparisons.
	Compared with the ImgHOID based methods HO-RCNN and iCAN, the proposed object trajectory detection module can effectively generate accurate trajectories for both individuals and objects, which provide temporally-consistent location information of the entities and help the proposed interaction recognition module to encode the dynamic human behavior and their relative motion.
	Compared with the VidVRD based methods VidVRD and STGCN, the proposed method focuses on analyzing fine-grained human behavior so as to pay more attention on human-centric visual content.
	Compared with VidHOIR based method TRN, our method generates instance-level HOI instances by directly analyzing upon specific human-object trajectory pairs.
	The higher tagging precision@1 and precision@5 also confirm that the capability of recognition can be improved with the assistance of explicit instance-level information.
	
	\section{Conclusion}
	
	In this paper, we proposed a new human-centric video understanding task named ST-HOID, which aims to describe the dominant visual content of videos with instance-level HOIs, and requires both spatial-temporal object localization and dynamic interaction recognition.
	We propose a novel ST-HOID method consisting of object trajectory detection and interaction reasoning, which aims to construct a new baseline framework for further research.
	To evaluate the effectiveness of the proposed method, we constructed a ST-HOID dataset.
	The experiment results show that the proposed method is superior to the state-of-the-art baselines.
	
	
	~\\
	\noindent\textbf{Acknowledgments} This work is supported by National Science Foundation of China (62072232), Natural Science Foundation of Jiangsu Province (BK20191248), Shenzhen Fundamental Research Program (JCYJ20180307151516166), and Collaborative Innovation Center of Novel Software Technology and Industrialization. 
	
	\bibliographystyle{IEEEbib}
	\bibliography{icme2021template}
	
\end{document}